\documentclass{article}
\usepackage{ijcai25}

\usepackage{times}
\usepackage{soul}
\usepackage{url}
\usepackage[hidelinks]{hyperref}
\usepackage[utf8]{inputenc}
\usepackage[small]{caption}
\usepackage{graphicx}
\usepackage{amsmath}
\usepackage{amsthm}
\usepackage{booktabs}
\usepackage{algorithm}
\usepackage{algorithmic}
\usepackage[switch]{lineno}

\usepackage{array}
\usepackage{graphicx}
\usepackage{tabularray}
\usepackage{multirow}
\usepackage{booktabs}
\usepackage{float}
\usepackage{placeins}
\usepackage{cleveref}
\usepackage{amssymb}
\usepackage{arydshln}
\usepackage{caption}

\title{\LARGE \textbf{Reliability-Driven LiDAR-Camera Fusion for Robust 3D Object Detection}}

\author{
\normalfont
\textit{Reza Sadeghian}$^1$, \textit{Niloofar Hooshyaripour}$^1$, \textit{Chris Joslin}$^2$, \textit{WonSook Lee}$^1$
\\
\vspace{3.5pt}
\affiliations
$^1$School of Electrical Engineering and Computer Science, University of Ottawa, Ottawa, Canada\\
$^2$School of Information Technology, Carleton University, Ottawa, Canada\\
\emails
\{r.sadeghian, nhoos082, wslee\}@uottawa.ca,
chris.joslin@carleton.ca
}
\begin{document}

\maketitle

\begin{abstract}
Accurate and robust 3D object detection is essential for autonomous driving, where fusing data from sensors like LiDAR and camera enhances detection accuracy. However, sensor malfunctions, such as corruption or disconnection, can degrade performance, and existing fusion models often struggle to maintain reliability when one modality fails. To address this, we propose ReliFusion, a novel LiDAR-camera fusion framework operating in the bird’s-eye-view (BEV) space. ReliFusion integrates three key components: the Spatio-Temporal Feature Aggregation (STFA) module, which captures dependencies across frames to stabilize predictions over time; the Reliability module, which assigns confidence scores to quantify the dependability of each modality under challenging conditions; and the Confidence-Weighted Mutual Cross-Attention (CW-MCA) module, which dynamically balances information from LiDAR and camera modalities based on these confidence scores. Experiments on the nuScenes dataset show that ReliFusion significantly outperforms state-of-the-art methods, achieving superior robustness and accuracy in scenarios with limited LiDAR fields of view and severe sensor malfunctions. %This framework sets a new benchmark for multimodal fusion in complex autonomous driving environments.

\end{abstract}

\section{Introduction}

\label{sec:intro}
Reliable 3D object detection is necessary in autonomous driving, where both LiDAR and camera sensors contribute complementary spatial and semantic data essential for precise environmental perception. While LiDAR captures detailed 3D spatial information, cameras provide rich semantic details, making multimodal fusion a promising solution for robust perception. Despite these advantages, many existing fusion approaches are sensitive to sensor malfunctions; when one modality experiences corruption—such as LiDAR occlusion or camera obstruction—it can severely degrade the performance of the other modality, compromising overall detection accuracy and robustness \cite{paden2016survey,tao2023pseudo,yu2023pipc}.

% \begin{figure}[htbp]
%     \centering
%     \begin{tblr}{
%   colspec = {X[3.5cm,c,h]X[4cm,c,h]},
%   stretch = 0,
%   rowsep = 0pt,
%   %hlines = {5, 1pt},
%   %vlines = {5, 1pt},
% }
%     %\cline{0-3}
%   \includegraphics[width=0.22\textwidth]{Images/pred_30 (1).png} &  \includegraphics[width=0.22\textwidth]{Images/pred_21 (1).png} \\

% \end{tblr}
% \caption{Left: Poor Illumination. Right: High Occlusion. Both conditions illustrate object detection challenges effectively addressed by our proposed network.}
% \label{fig:Condition}
% \end{figure}
\begin{figure}[ht]
    \centering
    \includegraphics[width=0.95\columnwidth]{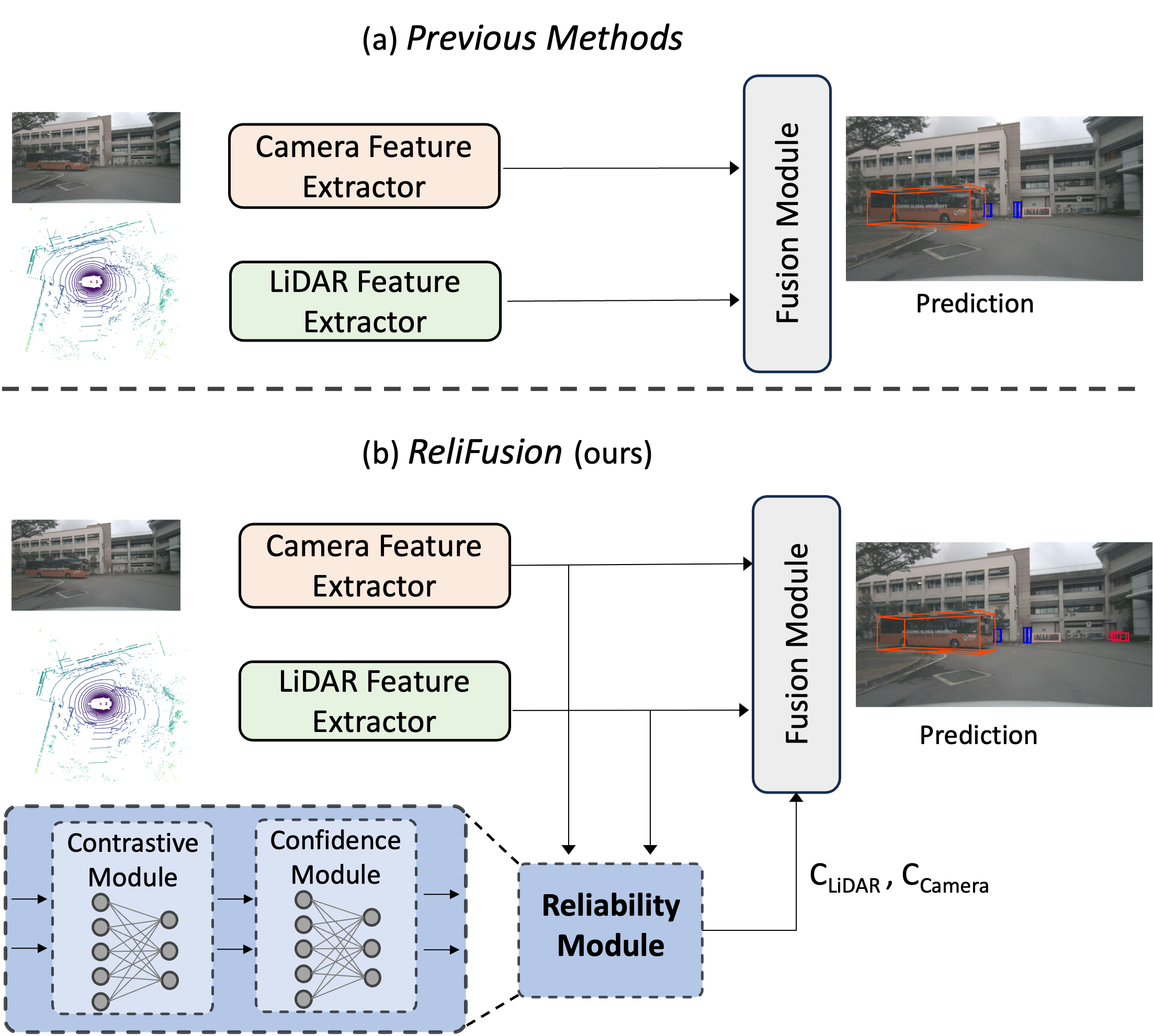} % Adjust width as needed
    
    \caption{Illustration of ReliFusion's approach compared to previous methods. (a) Traditional methods rely on fixed fusion mechanisms, struggling under sensor malfunctions. (b) ReliFusion introduces a Reliability Module, incorporating contrastive and confidence modules, to assign confidence scores for LiDAR ($C_{\text{LiDAR}}$) and camera ($C_{\text{Camera}}$). These scores enable dynamic balancing of LiDAR and camera contributions within the fusion module, achieving robust and accurate 3D object detection even under challenging conditions.}

    \label{fig:difference}
\end{figure}

Current fusion techniques often rely heavily on the LiDAR branch for accurate 3D localization, as seen in LiDAR-dominant models that use early, intermediate, or late fusion strategies \cite{sindagi2019mvx,wang2021pointaugmenting,chen2022autoalign}. However, this dependency poses a risk; if LiDAR data is partially compromised, it frequently leads to unreliable or even failed detections. Some models have attempted to address this by integrating camera data in a bird's-eye-view (BEV) space, where decoupled feature extraction enables each modality to be processed independently \cite{liang2022bevfusion,liu2023bevfusion,yu2023benchmarking}.

Nonetheless, these methods remain susceptible to single-modality failures due to their inability to adequately adjust fusion weights based on the reliability of the input from each sensor.

In response to the limitations of existing fusion models under sensor malfunctions, we propose ReliFusion, a novel LiDAR-camera fusion framework designed to enhance robustness in 3D object detection by dynamically adapting to varying sensor reliability. ReliFusion operates in the BEV space and integrates three key components. The STFA module processes the input first, capturing dependencies across frames to stabilize predictions over time and enhance detection consistency. Following this, the Reliability module leverages Cross-Modality Contrastive Learning (CMCL) to align LiDAR and camera features within a shared embedding space, distinguishing between reliable and corrupted data pairs and generating confidence scores that reflect each modality’s reliability. Finally, the CW-MCA module uses these confidence scores to dynamically fuse LiDAR and camera data, assigning higher weight to the more reliable modality and mitigating the impact of degraded inputs.

Our experiments on the nuScenes \cite{caesar2020nuscenes} dataset validate the effectiveness of ReliFusion, demonstrating significant improvements in detection accuracy and resilience under various sensor degradation scenarios. By addressing the challenges of sensor reliability and fusion, ReliFusion establishes a robust approach for accurate object detection in challenging conditions commonly encountered in autonomous driving.

Our primary contributions can be summarized as follows:
\begin{itemize}
    \item We develop a Reliability module that leverages CMCL to generate confidence scores, effectively quantifying the reliability of each modality and distinguishing between corrupted and reliable data. This module is critical in ensuring adaptive and robust multimodal fusion.

         \item We introduce a CW-MCA module that utilizes the confidence scores from the Reliability module to dynamically balance contributions from LiDAR and camera data, ensuring robust fusion even in degraded sensor conditions.
    
    % \item We incorporate a STFA module to capture spatial and temporal dependencies across frames, stabilizing predictions and improving detection consistency under transient malfunctions.
    
     \item We propose ReliFusion, a novel LiDAR-camera fusion framework that integrates these components to dynamically adapt to sensor reliability, enhancing robustness and accuracy for 3D object detection in challenging scenarios.

\end{itemize}
%By advancing the state-of-the-art in object detection through our contributions, we aim to pave the way for more robust and accurate systems that can handle complex real-world scenarios effectively. These findings hold significant implications for a wide range of applications, from autonomous vehicles to surveillance and robotics, where reliable object detection is of paramount importance.

The remainder of this paper is organized as follows: \Cref{sec:Related} provides an overview of related work in 3D object detection . \Cref{sec:TransfuseNet} provides a detailed explanation of ReliFusion, thoroughly describing its core modules and their interactions. The experimental results are presented in \Cref{sec:Experiment}, followed by the conclusion and future research directions in \Cref{sec:conclusion}.

\section{Related Work}
\label{sec:Related}

\subsection {Single-Modality 3D Object Detection}
Single-modality 3D object detection methods are based solely on either LiDAR or camera data, each providing specific advantages. LiDAR-based techniques leverage precise spatial data to generate accurate 3D representations. PointNet \cite{qi2017pointnet++} pioneered direct processing of point clouds, which was further developed by VoxelNet \cite{zhou2018voxelnet}, introducing voxelized features for improved efficiency and spatial detail. Subsequent methods such as SECOND \cite{yan2018second} and PV-RCNN \cite{shi2020pv} expanded on these by enhancing spatial representation and optimizing feature extraction, producing more reliable bounding box predictions. On the other hand, camera-only approaches, though limited in depth accuracy, provide semantic richness essential for object classification. DETR3D \cite{wang2022detr3d} uses transformers to lift 2D image features into 3D space, and BEVDepth \cite{li2023bevdepth} enhances depth estimation, achieving better 3D localization through refined view transformations. However, single-modality approaches inherently lack the complementary insights that multimodal fusion can offer.

\subsection{ Multimodal Sensor Fusion}
Multimodal fusion methods integrate spatially rich LiDAR data with semantically informative camera data, forming a comprehensive perception model. BEV fusion has become a common framework for such integration, with models like BEVFusion \cite{liang2022bevfusion,liu2023bevfusion} and related models \cite{jiao2023msmdfusion,li2024gafusion} adopting lift-splat-shoot (LSS) transformations \cite{philion2020lift} to align image data in BEV space, allowing it to be fused with LiDAR features effectively. This approach enables the model to capture both spatial geometry and semantic richness, creating a unified feature space that enhances detection performance. CMT \cite{yan2023cross} and MSMD-Fusion \cite{jiao2023msmdfusion} utilize attention mechanisms and hierarchical fusion strategies to align and integrate LiDAR and camera features. While CMT models interactions through transformers and MSMD-Fusion employs multiscale fusion, these methods do not account for sensor reliability, limiting performance under sensor degradation.

\subsection{Temporal Fusion}
Temporal fusion techniques aggregate features across multiple frames, improving detection performance by capturing motion and continuity. BEVDet4D \cite{huang2022bevdet4d} and BEVFormer \cite{li2022bevformer} leverage temporal BEV representations to consolidate information over time, enhancing robustness against transient occlusions. For example, BEVDet4D coordinates frame-by-frame BEV features, while BEVFormer applies spatio-temporal transformers to integrate cross-frame data, achieving stable detection over time. Similarly, 3D-VID \cite{zhai2022vid} utilizes attention mechanisms across point cloud frames to capture object transformations, offering improved detection in dynamic driving scenarios. Although temporal fusion captures scene continuity, it does not fully address issues of degraded data within individual frames.

\subsection{Robustness of LiDAR and Camera Fusion}
Ensuring robustness in LiDAR-camera fusion has become increasingly important, especially when dealing with noisy or partially corrupted data from either modality. TransFusion \cite{bai2022transfusion} uses transformer-based adaptive weighting to prioritize reliable sensor inputs, showing potential in managing modality-specific reliability. GAFusion \cite{li2024gafusion} refines this further, using LiDAR-derived depth information to guide adaptive fusion, selectively refining camera features to enhance cross-modal interaction under adverse conditions. SparseFusion \cite{xie2023sparsefusion} further enhances robustness by employing sparse representations from both modalities, increasing efficiency while managing data quality in challenging scenarios. Although these approaches improve the robustness of the fusion, they often lack explicit mechanisms for dynamically adjusting fusion weights based on real-time sensor reliability, leaving the fusion process vulnerable to reliability issues under adverse sensor conditions.

\begin{figure*}[tb]
    \centering
  \includegraphics[width=1\textwidth]{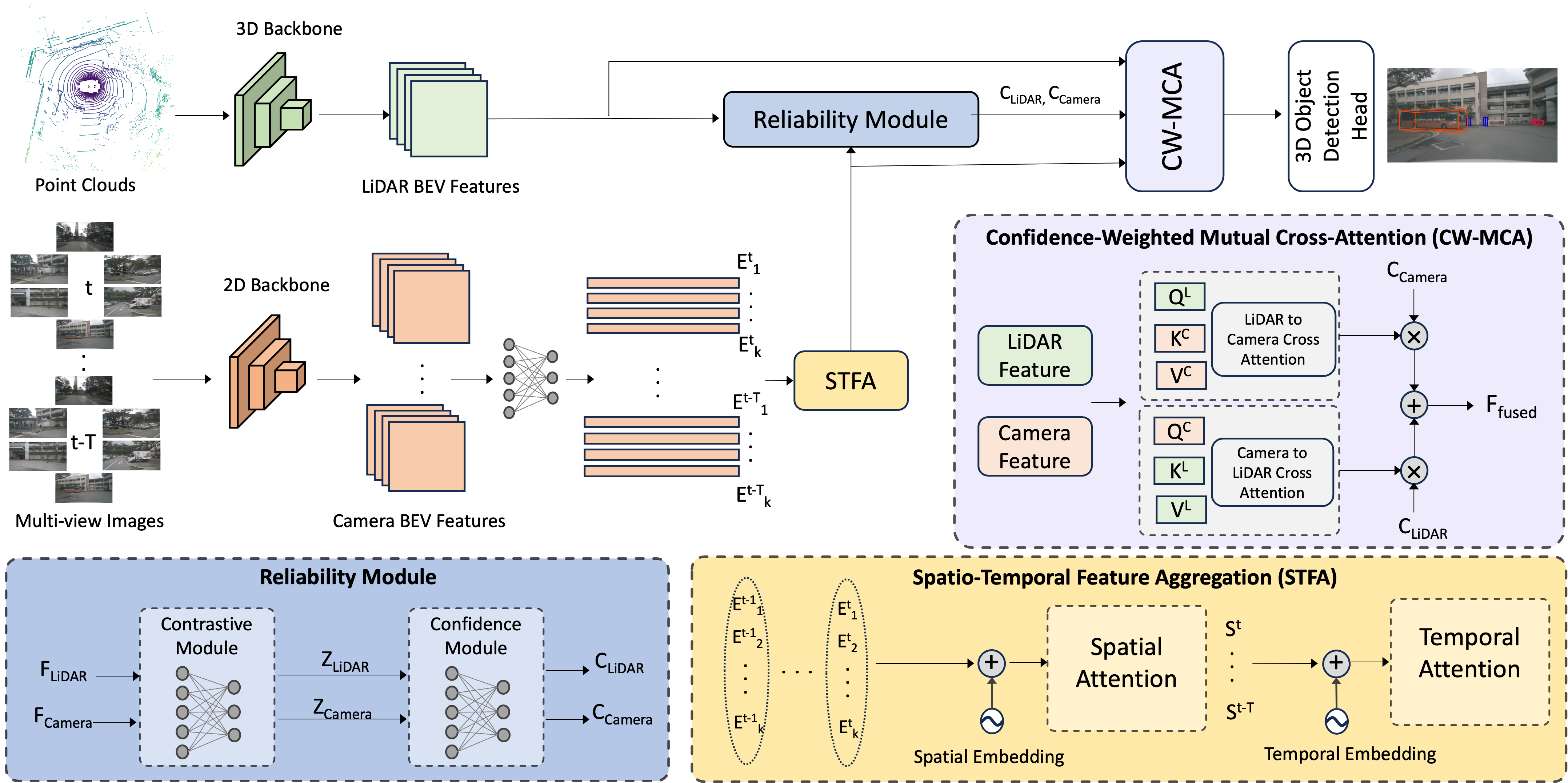}
  \caption{The overal architecture of ReliFusion.}
  
  \label{fig:Network_Reli}
\end{figure*}

ReliFusion addresses the limitations of existing methods by introducing a Reliability module that dynamically adjusts the contribution of LiDAR and camera features based on real-time confidence scores. These scores, derived through CMCL, guide the CW-MCA module for robust multimodal fusion. Additionally, a STFA module enhances detection stability by leveraging cross-frame dependencies, ensuring accurate performance even under sensor degradation.

\section{ReliFusion}
\label{sec:TransfuseNet}
Our proposed method integrates LiDAR point cloud and multi-view image data through a robust fusion framework, with each component pretrained independently to enhance specific functionalities before end-to-end integration. As shown in Figure \ref{fig:Network_Reli}, ReliFusion consists of five key components: (1) LiDAR and multi-view image feature extraction, 
%(2) BEV feature transformation for images, %
(2) STFA for temporal consistency, (3) Reliability Module for reliability assessment, and (4) CW-MCA for final fusion.

\subsection{Multi-View Image and LiDAR Feature Extraction}

The input to our model consists of LiDAR point clouds and synchronized multi-view images. The LiDAR point cloud is processed through a 3D backbone network, which generates spatially accurate BEV features, denoted as \( F_{\text{LiDAR}} \). Simultaneously, multi-view images are passed through a 2D convolutional backbone to produce initial image features (\( F_{\text{Camera}} \)) for each view.

To integrate these image features with the LiDAR-derived BEV features, we transform the image features into a unified BEV space using a lift-splat-shoot (LSS) operation \cite{philion2020lift}. This operation aggregates multi-view image data in BEV space, allowing them to be fused with the LiDAR BEV features. The resulting BEV-transformed image features \( F_{\text{Camera-BEV}} \) serve as a foundation for subsequent temporal and reliability-focused processing steps.

\subsection{Spatio-Temporal Feature Aggregation (STFA)}

Accurate 3D object detection in dynamic environments requires a robust representation that captures both spatial relationships within each time step and temporal dependencies across time steps. We introduce the STFA module, which sequentially applies spatial attention and temporal attention mechanisms. This module incorporates learnable spatial embeddings and temporal encodings to ensure effective feature aggregation.

\subsubsection{Spatial Attention for Inter-View Aggregation.}

At each time step \(t\), multi-view image features \(\{F^t_k\}_{k=1}^6\), where \(k\) indexes the views, are extracted using a shared CNN backbone. These BEV features \(F^t_k \in \mathbb{R}^{C \times H \times W}\) are flattened and projected into an embedding space:
\begin{equation}
E_k^t = W_s^\top \cdot \text{Flatten}(F_k^t) + b_s
\end{equation}
where \( W_s^\top \in \mathbb{R}^{d \times (C \cdot H \cdot W)} \) is a learnable weight matrix, \(b_s \in \mathbb{R}^d\) is a bias vector, and \(\text{Flatten}(\cdot)\) reshapes the feature map into a sequence.

To model relationships across views at the same time step, we employ a \textit{spatial self-attention mechanism}. For a given view \(k \in \{1, \dots, 6\}\) (query), the embeddings \(E_j^t\) from all views \(j \in \{1, \dots, 6\}\) (keys and values) are considered, including \(j = k\). The attention weights are computed as:
\begin{equation}
    \text{Attention}_s(Q_k, K_j) = \text{Softmax}\left(\frac{Q_k K_j^\top}{\sqrt{d}}\right)
\end{equation}
where:
\begin{equation}
    Q_k = W_q E_k^t , \quad K_j = W_k E_j^t , \quad V_j = W_v E_j^t
\end{equation}
and \(W_q, W_k, W_v \in \mathbb{R}^{d \times d}\) are learnable projection matrices.

The attended value for view \(k\) at time \(t\) is given by:
\begin{equation}
    S_k^t = \sum_{j=1}^6 \text{Attention}_s(Q_k, K_j) V_j
\end{equation}

Finally, the spatially aggregated features for all views are combined into a single representation for time \(t\):
\begin{equation}
    S^t = \{S_k^t \mid k = 1, \dots, 6\}
\end{equation}

This process ensures that each view attends to all others, effectively combining complementary information while retaining self-similarity (\(k = j\)). The spatial self-attention mitigates occlusions and enhances the spatial consistency of the aggregated features~\cite{liu2021swin,vaswani2017attention}.

\subsubsection{Temporal Attention for Cross-Time Dependency.}

The spatially aggregated features \(\{S^t\}_{t=1}^T\), where \(T\) denotes the number of time steps, are processed through a temporal attention mechanism to capture dependencies across time. Temporal embeddings are added to encode sequential order:
\begin{equation}
    \tilde{S}^t = S^t + P_t(t)
\end{equation}
where \(P_t(t) \in \mathbb{R}^d\) represents the learnable temporal encoding for each time step \(t\).

Temporal attention computes relationships across all time steps:
\begin{equation}
    \text{Attention}_t(Q, K, V) = \text{Softmax}\left(\frac{QK^\top}{\sqrt{d}}\right)V
\end{equation}
where:
\begin{equation}
    Q = W_q \tilde{S}^t, \quad K = W_k \tilde{S}^{t'}, \quad V = W_v \tilde{S}^{t'}
\end{equation}
for \(t' \neq t\), and \(W_q, W_k, W_v \in \mathbb{R}^{d \times d}\) are learnable projection matrices. The aggregated temporal feature is:
\begin{equation}
    T = \sum_{t=1}^T \text{Attention}_t(Q, K, V)
\end{equation}

This step models temporal correlations, ensuring that features across different time steps are effectively integrated. By capturing temporal dynamics, the temporal attention mechanism enhances the model's ability to handle motion and other time-dependent phenomena \cite{bertasius2021space,vaswani2017attention}.

\subsubsection{Refinement with Layer Normalization}

Each attention mechanism is followed by layer normalization and residual connections to stabilize training \cite{vaswani2017attention}:
\begin{equation}
    \hat{T} = \text{LayerNorm}(T + \text{MLP}(T))
\end{equation}
where \(\text{MLP}\) is a two-layer feedforward network with GELU activation \cite{hendrycks2016gaussian}. This refinement step ensures that the aggregated features are consistent and ready for downstream fusion. The output of the STFA module, \(\hat{T}\), serves as a spatiotemporally consistent feature representation which will be fused with LiDAR features. %It is fused with complementary modalities (e.g., LiDAR features) in the detection pipeline, enabling robust 3D object detection even in challenging scenarios with dynamic motion or occlusions.

\subsection{Reliability Module}
The temporally consistent BEV features  from the STFA module, along with the LiDAR features, are passed into the Reliability module, which leverages CMCL to align features from LiDAR and camera data while assessing their reliability. CMCL operates in two stages: feature alignment and reliability scoring.

During feature alignment, CMCL embeds LiDAR and image features into a shared embedding space (\( z_{\text{LiDAR}} \) and \( z_{\text{Camera-BEV}} \)) using multi-layer perceptrons (MLPs). Positive pairs (unaltered data) are encouraged to be close in the embedding space, while negative pairs (corrupted data) are pushed apart. This is achieved using a contrastive loss function:
\begin{equation}
L_{\text{contrast}} = -\log \frac{\exp(\text{sim}(z_{\text{LiDAR}}, z_{\text{Camera}}) / \tau)}{\sum_{i=1}^{K} \exp(\text{sim}(z_{\text{LiDAR}}, z_{\text{Camera}}^i) / \tau)}
\end{equation}
where \( \text{sim}(z_{\text{LiDAR}}, z_{\text{Camera}}) \) represents the cosine similarity between embeddings, and \( \tau \) is a temperature parameter, and K is the batch size.

%By aligning features across modalities, this process improves robustness and ensures effective feature integration, even under varying data quality conditions. Pre-training the contrastive module allows it to produce well-aligned embeddings \( z_{\text{LiDAR}} \) and \( z_{\text{Image-BEV}} \), making them suitable for reliable cross-modality coherence.

%\subsection{Confidence Module}
In the second stage, CMCL generates reliability scores for each modality (\( C_{\text{LiDAR}} \) and \( C_{\text{Camera}} \)) by analyzing the quality of their embeddings. These scores quantify the confidence in each modality’s features and are computed as follows:
\begin{equation}
\begin{aligned}
C_{\text{LiDAR}} &= \sigma\big(W_{\text{LiDAR}} z_{\text{LiDAR}} + b_{\text{LiDAR}}\big) \\
C_{\text{Camera}} &= \sigma\big(W_{\text{Camera}} z_{\text{Camera}} + b_{\text{Camera}}\big)
\end{aligned}
\end{equation}
where \( W \) and \( b \) are learnable parameters, and \( \sigma \) denotes the sigmoid activation function. These scores enable dynamic weighting of the contributions from each modality during the fusion process.

\begin{table*}[h]

\begin{center}
\resizebox{\textwidth}{!}{%
\setlength{\extrarowheight}{2pt}
\begin{tabular}{l|cc|cc|cccccccccc}
\hline 

%\cline{2-4} 
 Method&L& C&mAP&NDS&Car& Truck& C.V.& Bus& Trailer& Barrier& Motor.& Bike& Ped.& T.C.\\
\hline 

 BEVFormer\cite{li2022bevformer}& -& \checkmark&48.1& 56.9& 67.7& 39.2& 22.9& 35.7&39.6&62.5&47.9&40.7&54.4&70.3\\
 BEVDet\cite{huang2022bevdet4d}&-& \checkmark&42.4& 48.8& 64.3& 35.0& 16.2& 35.8&35.4&61.4&44.8&29.6&41.1&60.1\\
 DETR3D\cite{wang2022detr3d}& -& \checkmark&41.2& 47.9& 60.3& 33.3& 17.0& 29.0&35.8&56.5&41.3&30.8&45.5&62.7\\
 \hdashline
  CenterPoint\cite{yin2021center}& \checkmark& -&60.3& 67.3& 85.2& 53.5& 20.0& 63.6&56.0&71.1&59.5&30.7&84.6&78.4\\
   TransFusion-L\cite{bai2022transfusion}& \checkmark& -&65.5& 70.2& 86.2& 56.7& 28.2& 66.3&58.8&78.2&68.3&44.2&86.1&82.0\\
   \hdashline
    FUTR3D\cite{chen2023futr3d}& \checkmark& \checkmark&64.2& 68.0& 86.3& 61.5& 26.0& 71.9&42.1&64.4&73.6&\textbf{63.3}&82.6&70.1\\
 PointAugmenting\cite{wang2021pointaugmenting}& \checkmark& \checkmark&66.8& 71.0& 87.5& 57.3& 28.0& 65.2&60.7&72.6&74.3&50.9&87.9&83.6\\

 TransFusion\cite{bai2022transfusion}& \checkmark& \checkmark&68.9& 71.7& 87.1& 60.0& 33.1& 68.3&60.8&78.1&73.6&52.9&88.4&\textbf{86.7}\\
 BEVfusion\cite{liang2022bevfusion}& \checkmark& \checkmark&69.2& 71.8& 88.1& 60.9& 34.4& 69.3&62.1&\textbf{78.2}&72.2&52.2&89.2&85.2\\

CMT\cite{yan2023cross}&$\checkmark$&\checkmark&70.4& 73.0& 87.2& 61.5&\textbf{37.5}&72.4&62.8&74.7&79.4&58.3&86.9&83.2\\

      \hline
      \textbf{ReliFusion (ours)}  &$\checkmark$&$\checkmark$&\textbf{70.6}&\textbf{73.2}& \textbf{88.1}& \textbf{61.7}&35.9 &\textbf{72.9} &\textbf{63.0}&75.8&\textbf{79.5}&57.4&\textbf{89.3}&83.5  \\

\hline

\end{tabular}%
}

\end{center}
\caption{Evaluation results on nuScenes dataset. We evaluated ReliFusion against the  SOTA results on the test set. `L' and `C' represents LiDAR and Camera, respectively. `C.V', `Ped', and `T.C' stand for construction vehicle, pedestrian, and traffic cone, respectively.
 The best results appear in bold. }
\label{tab:comparison}
\end{table*}

% Based on the aligned embeddings \( z_{\text{LiDAR}} \) and \( z_{\text{Image-BEV}} \) from the Contrastive Module, the Confidence Module generates a confidence score for each modality, reflecting the reliability of its features for the current scene. These confidence scores, \( C_{\text{LiDAR}} \) and \( C_{\text{Image}} \), are produced by applying a sigmoid-activated MLP to each embedding:
% \begin{equation}
% C_{\text{LiDAR}} = \sigma(\text{MLP}_{\text{LiDAR}}^{\text{conf}}(z_{\text{LiDAR}})), \quad C_{\text{Image}} = \sigma(\text{MLP}_{\text{Image}}^{\text{conf}}(z_{\text{Image}})),
% \end{equation}
% where \( \sigma \) denotes the sigmoid activation function.

%These confidence scores range from 0 to 1, providing a dynamic reliability assessment for each modality that will guide the final fusion process. Pretraining the Confidence Module enables it to accurately assess modality reliability, yielding reliable confidence scores \( C_{\text{LiDAR}} \) and \( C_{\text{Image}} \) that reflect scene-specific conditions.

\subsection{Confidence-Weighted Mutual Cross-Attention (CW-MCA)}

The CW-MCA module performs the final fusion of  \(F_{\text{LiDAR}}\) and \(F_{\text{Camera-BEV}}\) using confidence scores derived from the Reliability module. %These scores dynamically adjust the contributions of each modality, ensuring robust fusion under varying sensor conditions.
% CW-MCA incorporates reliability scores \(C_{\text{LiDAR}}\) and \(C_{\text{Camera}}\) to modulate cross-modal interactions. 
The confidence-weighted feature representations are computed as:
\begin{equation}
F_{\text{L} \rightarrow \text{C}} = C_{\text{LiDAR}} \cdot \text{softmax}\left(\frac{Q_{\text{C}} K_{\text{L}}^\top}{\sqrt{d_k}}\right)V_{\text{L}}
\end{equation}
\begin{equation}
F_{\text{C} \rightarrow \text{L}} = C_{\text{Camera}} \cdot \text{softmax}\left(\frac{Q_{\text{L}} K_{\text{C}}^\top}{\sqrt{d_k}}\right)V_{\text{C}}
\end{equation}

The final fused feature representation is computed as:
\begin{equation}
F_{\text{fused}} = F_{\text{L} \rightarrow \text{C}} + F_{\text{C} \rightarrow \text{L}}
\end{equation}

By leveraging confidence scores, CW-MCA dynamically adapts to sensor reliability, ensuring accurate fusion and robust performance in degraded conditions.
Finally, we integrate the widely adopted TransFusion-head \cite{bai2022transfusion} into our framework to process \(F_{\text{fused}}\), enabling the prediction of object categories and regression of key attributes such as object dimensions, orientation, and velocity.

\subsection{End-to-End Fine-Tuning}

After pretraining each module, we fine-tune the entire network end-to-end to optimize for the object detection task. This stage uses a multi-task loss function that combines detection, contrastive, confidence, and temporal consistency losses:
\begin{equation}
\mathcal{L}_{\text{T}} = \lambda_{\text{1}} \mathcal{L}_{\text{det}} + \lambda_{\text{2}} 
\mathcal{L}_{\text{contrast}} + \lambda_{\text{3}} \mathcal{L}_{\text{temp}} + \lambda_{\text{4}} \mathcal{L}_{\text{conf}}
\end{equation}

where \( \mathcal{L}_{\text{det}} \) is the object detection loss, \( \mathcal{L}_{\text{contrast}} \) is the contrastive loss for cross-modality feature alignment, \( \mathcal{L}_{\text{temp}} \) ensures temporal consistency, and \( \mathcal{L}_{\text{conf}} \) penalizes inaccurate confidence predictions. $\lambda_{\text{1}}$, $\lambda_{\text{2}}$, $\lambda_{\text{3}}$, and $\lambda_{\text{4}}$ are the coefficients of the individual costs, which are set to 1.0, 0.1, 0.2, and 0.05, respectively.

\renewcommand{\arraystretch}{1.0}

\renewcommand{\arraystretch}{1}

\section{Experiments}
\label{sec:Experiment}
% In this section, we first describe the experimental setup. Then, evaluations are performed on the KITTI 2D/BEV object detection benchmark, considering various input modalities and fusion strategies for the late fusion stage. Moreover, Comparative analyses with state-of-the-art methods are provided, followed by ablation studies to validate design decisions. Finally, qualitative results and discussions are presented.
\subsection{Dataset and Metric}
We evaluate our method on the nuScenes dataset \cite{caesar2020nuscenes}, which includes synchronized data from six cameras covering 360-degree views and a LiDAR sensor. The dataset contains 1.4 million 3D bounding boxes across ten classes. Evaluation metrics include mean Average Precision (mAP) in BEV space and nuScenes Detection Score (NDS), which combines mAP with attributes like translation, scale, and orientation.

\subsection{Implementation details}

ReliFusion is implemented in PyTorch using the MMDetection3D framework \cite{mmdetection}. Following the approach in \cite{sadeghian2024transformer}, ConvMixer \cite{convmixer} is utilized as the 2D backbone, and VoxelNet \cite{zhou2018voxelnet} serves as the 3D backbone, with both projected into the BEV space. In line with CenterPoint \cite{yin2021center}, LiDAR point clouds are voxelized with a size of \(0.075 \, \text{m} \times 0.075 \, \text{m} \times 0.2 \, \text{m}\), and input images are resized to \(448 \times 800\) pixels.

Our training process consists of three stages. The first stage involves pre-training the Contrastive Module to align LiDAR and camera embeddings using a contrastive loss (temperature \(0.07\)) with an embedding size of 128, pre-training the Confidence Module to assign reliability scores via regression loss, and pre-training the STFA module to ensure temporal consistency across frames. In the second stage, the image stream and LiDAR stream are trained separately for 15 epochs to optimize their respective backbones and feature extraction processes. In the third and final stage, the entire model is fine-tuned end-to-end with a batch size of 16, an initial learning rate of \(1 \times 10^{-4}\), and the Adam optimizer (weight decay \(1 \times 10^{-5}\)) for 5 epochs. 
% The total loss function is:
% \begin{equation}
% \mathcal{L}_{\text{total}} = \lambda_{\text{det}} \mathcal{L}_{\text{det}} + \lambda_{\text{contrast}} \mathcal{L}_{\text{contrast}} + \lambda_{\text{conf}} \mathcal{L}_{\text{conf}} + \lambda_{\text{temporal}} \mathcal{L}_{\text{temporal}}
% \end{equation}
% where \(\lambda_{\text{det}} = 1.0\), \(\lambda_{\text{contrast}} = 0.1\), \(\lambda_{\text{conf}} = 0.05\), and \(\lambda_{\text{temporal}} = 0.2\).
%  This structured multi-stage training approach ensures each component is optimized for its specific task while maintaining robust and accurate multimodal fusion.

\subsection{State-of-the-Art Comparison}

Although the primary objective of ReliFusion is to improve robustness under challenging conditions, it also demonstrates superior performance on clean datasets. As shown in Table \ref{tab:comparison}, ReliFusion achieves an mAP of 70.6\% and an NDS of 73.2\%, outperforming existing SOTA methods such as BEVFusion \cite{liang2022bevfusion} and TransFusion \cite{bai2022transfusion}.

\begin{figure*}[tb]
    \centering
  \includegraphics[width=0.9\textwidth]{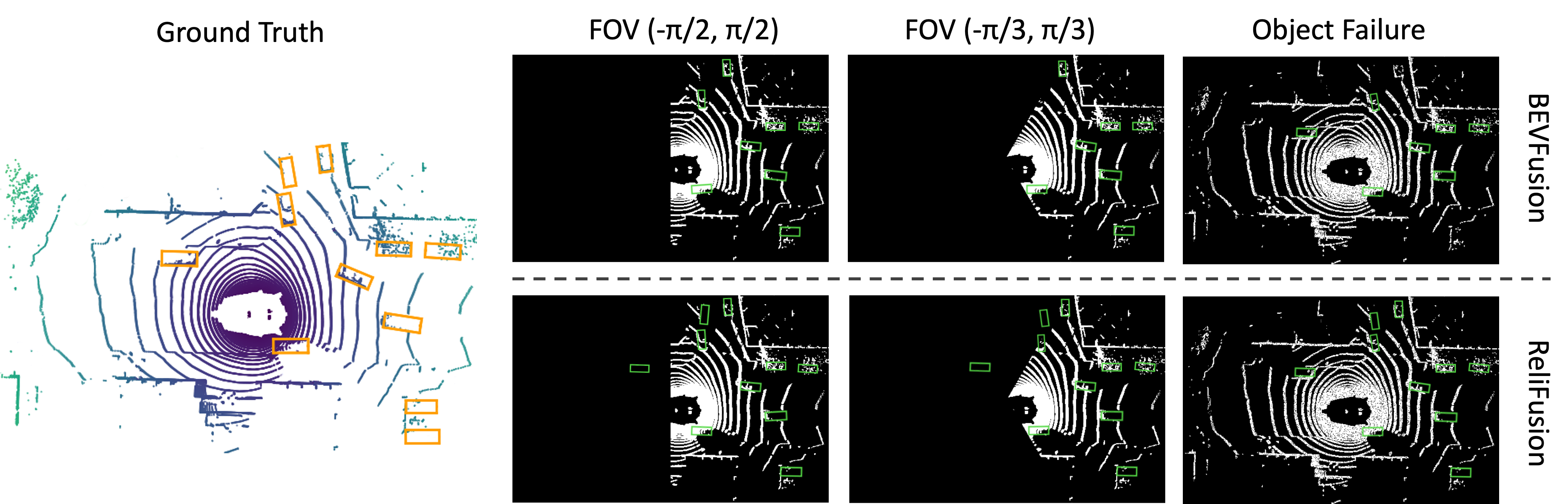}
  \caption{Qualitative detection results of BEVFusion and ReliFusion under LiDAR malfunctions scenarios. Clearly, BEVFusion struggles when LiDAR input is unavailable, whereas ReliFusion relies on camera to compensate and detect these objects. Green and Orange bounding boxes are true positive detection and ground truth, respectively.}
  
  \label{fig:Network}
\end{figure*}

\begin{table*}[htbp]
\centering
\setlength{\extrarowheight}{1.3pt} % Add extra vertical space for all rows
\setlength{\tabcolsep}{5pt} % Adjust column spacing
\fontsize{9pt}{11pt}\selectfont % Adjust font size
\begin{tabular}{l|c|c|c|c|c|c}
\hline
\multirow{2}{*}{Method}        & \multirow{2}{*}{Modality} & \multirow{2}{*}{Clean} & \multicolumn{3}{c|}{Limited LiDAR FOV} & LiDAR Object Failure \\ \cline{4-6}
                       &                  &                & ($-\pi/2, \pi/2$) & ($-\pi/3, \pi/3$) & ($-0, 0$) & 50\% Drop \\ \hline
CenterPoint\cite{yin2021center}      & L                & 56.8/65.0      & 23.5/47.7 & 15.6/43.0 & 0/0            & 28.4/48.5   \\ 
PointAugmenting\cite{wang2021pointaugmenting}   & LC               & 46.9/55.6      & 19.5/41.2 & 13.3/37.7 & 0/0            & 21.3/39.4   \\ 
MVX-Net \cite{sindagi2019mvx}           & LC               & 61.0/66.1      & 26.0/47.8 & 17.6/43.1 & 0/0            & 34.0/51.1   \\ 
TransFusion\cite{bai2022transfusion}       & LC               & 66.9/70.9      & 29.3/51.4 & 20.3/45.8 & 0/0            & 34.6/53.6   \\ 
BEVFusion\cite{liang2022bevfusion}          & LC               & 67.9/71.0      & 46.4/55.8 & 41.5/50.8 & 12.4/17.1      & 50.3/57.6   \\ 
\textbf{ReliFusion (ours)} & LC               & \textbf{70.6/73.2} & \textbf{52.4/59.6} & \textbf{44.9/54.6} & \textbf{24.6/28.7} & \textbf{53.1/60.6} \\ \hline
\end{tabular}%

\caption{Comparison of SOTA methods under limited LiDAR FOV and object failure scenarios, with mAP and NDS metrics provided.}
\label{tab:lidarMAl}
\end{table*}

\subsection{Robustness Experiments}

\subsubsection{Robustness Experiments against LiDAR Malfunctions.}

To evaluate the resilience of ReliFusion, we conducted experiments simulating various LiDAR malfunctions. These tests were designed to examine the model's capacity to sustain accurate object detection under conditions of degraded or reduced LiDAR inputs. Two key types of LiDAR malfunctions were simulated:

\begin{itemize}
\item \textbf{Limited Field of View (FOV):} LiDAR points were retained within progressively narrower angular ranges, such as ($-\pi/2, \pi/2$), ($-\pi/3, \pi/3$), and ($-0, 0$), reflecting practical challenges like occlusions and hardware limitations.
\item \textbf{Object Detection Failures:} Randomized point drops were applied within object bounding boxes at a rate of 50\%, simulating scenarios such as poor surface reflectivity or severe sensor degradation.
\end{itemize}

The findings, summarized in Table~\ref{tab:lidarMAl}, highlight the robustness of ReliFusion compared to state-of-the-art methods. For Limited FOV, LiDAR-only models such as CenterPoint \cite{yin2021center} exhibited a steep decline in performance due to their complete reliance on LiDAR data, while non-BEV methods like TransFusion \cite{bai2022transfusion} also suffered significant drops in mAP and NDS because they heavily depend on LiDAR features for region proposals or key points, using image data only as auxiliary input. Although BEV-based methods attempt to maximize the utilization of camera information by fusing data within the same spatial domain, the contribution from cameras alone often falls short of achieving robust results. For instance, models like BEVFusion \cite{liang2022bevfusion} address some of these challenges by leveraging complementary information from camera inputs, leading to improved mAP scores of 46.4\%, 41.5\%, and 50.3\% in various scenarios. However, these methods do not fully account for the dynamic reliability of each modality, especially under sensor malfunctions. ReliFusion advances this by determining the appropriate contribution of each modality, enhancing overall robustness in the presence of sensor malfunctions. This robustness becomes particularly evident when LiDAR data is unavailable, underscoring ReliFusion's effectiveness as a SOTA approach in such conditions.

%Our proposed \textit{ReliFusion} further enhances robustness by integrating confidence-weighted fusion and dynamic cross-modal feature modeling. Specifically, \textit{ReliFusion} achieved 48.4\% mAP and 58.2\% NDS for (), 42.9\% mAP and 53.6\% NDS for (), and 23.6\% mAP and 28.7\% NDS for (), outperforming all competing approaches. For \textbf{Object Detection Failures}, where 50\% of LiDAR points within object bounding boxes were dropped, \textit{ReliFusion} attained 53.1\% mAP and 60.6\% NDS. The framework’s decoupled feature extraction and holistic BEV representation allowed it to adapt to corrupted inputs by prioritizing reliable camera features, achieving significant performance gains over non-BEV and LiDAR-dominant models.

%These results emphasize the critical importance of dynamic adaptation to sensor reliability. By leveraging complementary camera information, temporal cues, and confidence-weighted fusion, ReliFusion delivers robust detection performance even under severe LiDAR malfunctions, setting a new standard in multimodal fusion frameworks.

\begin{table*}[htbp]

\centering
\setlength{\extrarowheight}{1.3pt} % Add extra vertical space for all rows
\setlength{\tabcolsep}{5pt} % Adjust column spacing
\fontsize{9pt}{11pt}\selectfont % Adjust font size

\begin{tabular}{l|c|c|c|c|c}
\hline
\multirow{2}{*}{Method}          &  \multirow{2}{*}{Modality} & \multirow{2}{*}{Clean} & \multicolumn{2}{c|}{Object Failure} & Object Occlusion \\ \cline{4-5}
                       &                  &                & Missing F & Preserve F & 50\% Occlusion \\ \hline

DETR3D\cite{wang2022detr3d}                      & C                & 34.9/43.4      & 25.8/39.2 & 3.3/20.5             & 14.3/29.0   \\ 
PointAugmenting \cite{wang2021pointaugmenting}   & LC               & 46.9/55.6      & 42.4/53.0 & 31.6/46.5             & 40.7/52.2   \\ 
MVX-Net\cite{sindagi2019mvx}                     & LC               & 61.0/66.1      & 47.8.0/59.4 & 17.5/41.7             & 45.5/57.6   \\ 
TransFusion \cite{bai2022transfusion}            & LC               & 66.9/70.9      & 65.3/70.1 & 64.4/69.3             & 65.5/70.0   \\ 
BEVFusion \cite{liang2022bevfusion}              & LC               & 67.9/71.0      & 65.9/70.7 & 65.1/69.9             & 65.9/70.1   \\ 
\textbf{ReliFusion (ours)} & LC                 & \textbf{70.6/73.2} & \textbf{68.3/71.3} & \textbf{65.9/70.4} & \textbf{67.8/70.6} \\ \hline

\end{tabular}

\caption{Comparison of SOTA methods under camera failure and object occlusion scenarios, with mAP and NDS metrics provided.}
\label{tab:CamMal}
\end{table*}

\begin{table*}[htbp]
\centering
\setlength{\extrarowheight}{1.3pt} % Add extra vertical space for all rows
\setlength{\tabcolsep}{5pt} % Adjust column spacing
\fontsize{9pt}{11pt}\selectfont % Adjust font size
\begin{tabular}{c@{\hspace{11pt}}c@{\hspace{11pt}}c|c|c|c|c}
\hline
\multicolumn{3}{c|}{\textbf{ReliFusion Modules}} & \multicolumn{3}{c|}{\textbf{Limited LiDAR FOV}} & \textbf{LiDAR Object Failure} \\ \cline{4-6}
STFA & CW-MCA & Reliability & ($-\pi/2, \pi/2$) & ($-\pi/3, \pi/3$) & ($0, 0$) & 50\% Drop \\ \hline
- & - & - & 33.3/44.2 & 25.4/37.4 & 5.1/22.3 & 38.6/49.8 \\ 
\checkmark & - & - & 43.9/52.5 & 36.6/45.7 & 17.2/29.2 & 45.4/55.6 \\ 
\checkmark & \checkmark & - & 49.4/58.3 & 40.3/50.8 & 20.4/36.6 & 50.3/57.2 \\ 
\checkmark & \checkmark & \checkmark & \textbf{52.4/59.6} & \textbf{44.9/54.8} & \textbf{24.6/39.7} & \textbf{53.1/60.6} \\ \hline
\end{tabular}
\caption{Ablation study of ReliFusion components (STFA, CW-MCA, and reliability modules) under limited LiDAR FOV and object failure scenarios, with mAP and NDS metrics reported.}
\label{tab:ablation_mainComponents}
\end{table*}

\subsubsection{Robustness Experiments against Camera malfunctions.}

To evaluate the robustness of ReliFusion under camera malfunction scenarios, we followed the setup in \cite{yu2023benchmarking}. Two types of malfunctions were tested: Camera Failure, where the front (F) image was either entirely removed or preserved, and Object Occlusion, where 50\% of the image pixels within object bounding boxes were masked to simulate partial occlusion. These scenarios mimic real-world challenges such as sensor disconnection, extreme weather conditions, or obstacles partially blocking the camera's view.

As shown in Table \ref{tab:CamMal}, ReliFusion achieved 68.3\% mAP when the front image was missing and 65.9\% mAP when it was preserved under camera failure. For 50\% occlusion, it reached 67.8\% mAP, outperforming state-of-the-art methods with minimal performance drop. While camera malfunctions had a smaller impact compared to LiDAR malfunctions, as fusion methods primarily rely on LiDAR, ReliFusion demonstrated superior robustness, maintaining reliable performance in all scenarios.

%\subsection{Results} 

%As \Cref{tab:inputmodality} shows, we evaluated the performance using various combinations of input modalities and late fusion operators, which will be analyzed in the following. The baseline uses solely RGB data, excluding LiDAR. Through extensive testing, LiDAR's inclusion consistently boosts accuracy across categories, underlining its pivotal role in 2D object detection. 

% \begin{figure}[tb]
%     \centering

%   \includegraphics[width=0.8\textwidth]{Images/prednewcolor_10 copyfinalend2.png}\\   
%   % \includegraphics[width=8cm\textwidth]{Images/pred_10 copy.png} \\

% \caption{Sample from the KITTI dataset illustrating the capability of our network to accurately detect an object despite incorrect annotation. The green bounding box indicates True Positive detection, while the blue bounding box represents Ground Truth.}
% \label{fig:Qresult2} %in chie? dorsotesh mikonam
% \end{figure}

% \begin{table}[tb]
% \centering
% \caption{Evaluating the impact of the Spatio Temporal Feature Aggregation Module (STFA).}
% \begin{tabular}{c@{\hspace{11pt}}|c@{\hspace{15pt}}c@{\hspace{11pt}}c@{\hspace{11pt}}c@{\hspace{11pt}}}
% \hline
% Evaluation metric & \hspace{10pt}without STFA & Spatio & Temporal & Spatio-Temporal \\ \hline
% mAP$\uparrow$ & \hspace{10pt}67.2 & 69.3 & 68.9 & \textbf{70.6} \\ \hline
% NDS$\uparrow$ & \hspace{10pt}71.3 & 72.8 & 72.6 & \textbf{73.2} \\ \hline
% \end{tabular}
% \label{tab:STFA}
% \end{table}

\begin{table}[tb]
\centering
\setlength{\extrarowheight}{1.2pt} % Add extra vertical space for all rows
\setlength{\tabcolsep}{4pt} % Adjust column spacing
\fontsize{9pt}{11pt}\selectfont % Adjust font size

\begin{tabular}{l@{\hspace{11pt}}|c@{\hspace{11pt}}c@{\hspace{11pt}}}
\hline 
 Aggregation Method& mAP$\uparrow$ & NDS$\uparrow$ \\ \hline 
without STFA & 67.2 & 71.3 \\ 
Spatio & 69.3 & 72.8 \\ 
Temporal & 68.9 & 72.6 \\ 
Spatio-Temporal & \textbf{70.6} & \textbf{73.2} \\ \hline
\end{tabular}

\caption{Evaluating the impact of the STFA.}
\label{tab:STFA}
\end{table}

% \begin{table}[tb]
% \centering
% \caption{Evaluating the impact of the confidence-weighted mutual cross-attention (CW-MCA). }
% \begin{tabular}{c@{\hspace{11pt}}|c@{\hspace{11pt}}c@{\hspace{11pt}}c@{\hspace{11pt}}c@{\hspace{11pt}}c@{\hspace{11pt}}c@{\hspace{11pt}}}
% \hline
% Evaluation metric & \hspace{10pt}Add  & Cross/Image & Cross/LiDar &MCA &CW-MCA\\ \hline
% mAP$\uparrow$ & \hspace{10pt}65.1  & 66.6 &67.4 & 68.3 &\textbf{70.6} \\ \hline
% NDS$\uparrow$ &\hspace{10pt}68.7&70.1& 71.3 &72.5&\textbf{73.2}\\ \hline
% \end{tabular}
% \label{tab:CWMCA}
% \end{table}

\begin{table}[tb]
\centering
\setlength{\extrarowheight}{1pt} % Add extra vertical space for all rows
\setlength{\tabcolsep}{5pt} % Adjust column spacing
\fontsize{9pt}{11pt}\selectfont % Adjust font size

\begin{tabular}{l@{\hspace{11pt}}|c@{\hspace{11pt}}c@{\hspace{11pt}}}

\hline 
 Fusion Method & mAP$\uparrow$ & NDS$\uparrow$ \\ \hline 
Add & 65.1 & 68.7 \\ 
Cross/Image & 66.6 & 70.1 \\ 
Cross/LiDar & 67.4 & 71.3 \\ 
MCA & 68.3 & 72.5 \\ 
CW-MCA & \textbf{70.6} & \textbf{73.2} \\ \hline
\end{tabular}

\caption{Evaluating the impact of the CW-MCA.}
\label{tab:CWMCA}
\end{table}

\newcommand{\column}[2]{%
  % #1 = column of images
  % #2 = caption
  \begin{tabular}[b]{@{}c@{}}#1\\#2\end{tabular}%
}

% \begin{figure*}[htp]
% \centering

% \begin{tabular}{
%   @{}
%   *{2}{c@{\hspace{4pt}}} % adjust to your needs (one less than the total)s
%   c
%   @{}
% }

% \column{
%   \textbf{Ground truth}\\
%   \textbf{best result}\\
%   \textbf{Second best result}
% }{GS04}
% &
% \column{
%   \includegraphics[width=2.5cm]{Images/1.jpeg}\\
%   \includegraphics[width=2.5cm]{Images/1.jpeg}\\
%   \includegraphics[width=2.5cm]{Images/1.jpeg}
% }{NC05}
% &
% \column{
%   \includegraphics[width=2.5cm]{Images/1.jpeg}\\
%   \includegraphics[width=2.5cm]{Images/1.jpeg}\\
%   \includegraphics[width=2.5cm]{Images/1.jpeg}
% }{TP09}
% \end{tabular}

% \caption{Some images}

% \end{figure*}

% \begin{figure*}[htp]
% \documentclass{article}
% \usepackage{array}

% \begin{center}
% \begin{tabular}{ |c| m{1cm}| m{1cm} | } 
%   \hline
%   cell1 dummy text dummy text dummy text& \includegraphics[width=2.5cm]{Images/1.jpeg}& cell3 \\ 
%   \hline
%   cell1 dummy text dummy text dummy text & cell5 & cell6 \\ 
%   \hline
%   cell7 & cell8 & cell9 \\ 
%   \hline
% \end{tabular}
% \end{center}

% \end{figure*}

%%%%%

%\setlength{\parindent}{0pt}

%%%%

\subsection{Ablation Study}
 We perform ablation studies on the key components of ReliFusion, including the STFA, CW-MCA, and Reliability modules. We first evaluate the efficacy of the STFA module and its impact on the final results. As shown in Table 5, incorporating spatio-temporal attention significantly enhances performance. To analyze its components, we trained the model without STFA, with only spatial attention, and with only temporal attention.

The results in Table \ref{tab:STFA} reveal that spatial attention provides more meaningful contributions compared to temporal attention when used independently. This can be attributed to the fact that spatial information captures rich geometric and structural features of the scene, which are critical for object detection. Temporal attention, on the other hand, primarily relies on temporal correlations, which may provide limited benefits in scenarios with less dynamic changes or fewer temporal dependencies. When both are combined in the STFA module, the network effectively integrates spatial and temporal cues, leveraging complementary information to achieve superior performance. This enables STFA to improve model robustness and accuracy by effectively capturing static spatial details and dynamic temporal variations.

As shown in Table \ref{tab:CWMCA}, LiDAR as the query outperforms image as the query because LiDAR provides richer spatial and geometric information crucial for precise object localization, while image data is more susceptible to challenges like occlusion and lighting variations. Mutual cross-attention (MCA), which queries both LiDAR and image modalities, builds upon this by leveraging complementary features from both, leading to better performance compared to relying on a single modality. Finally, CW-MCA enhances MCA by dynamically weighting each modality based on its reliability. This confidence-weighted mechanism allows the model to adapt to varying sensor conditions, ensuring robust fusion and achieving the highest detection accuracy and robustness.

Finally, as shown in Table \ref{tab:ablation_mainComponents}, we ablated the main modules of ReliFusion, including STFA, CW-MCA, and the Reliability module, to evaluate their individual contributions. The results clearly demonstrate the efficacy of each component. The inclusion of STFA improves performance by leveraging temporal information from previous frames, allowing the model to rely more effectively on the image branch for better predictions. CW-MCA enhances the model's ability to fuse LiDAR and camera information dynamically, utilizing the complementary strengths of both modalities. 
% The Reliability module, the most critical part of our network, further improves performance by focusing on the modality with higher reliability, ensuring robust predictions even in challenging scenarios.

Notably, the impact of these modules becomes more pronounced as LiDAR limitations increase. For example, when LiDAR is completely corrupted, the model achieves 24.6\% mAP using only image information, a significant improvement compared to configurations without these modules. 

\section{Conclusion}
\label{sec:conclusion}
In this paper, we introduced ReliFusion, a novel LiDAR-camera fusion framework designed to enhance the robustness of 3D object detection in autonomous driving scenarios. By integrating STFA, Reliability module, and CW-MCA, ReliFusion effectively adapts to sensor degradation, dynamically prioritizing more reliable modalities. Our experiments on the nuScenes dataset demonstrated that our framework outperforms state-of-the-art methods across various challenging conditions, including limited LiDAR fields of view and severe sensor malfunctions. These results highlight the efficacy of our approach in leveraging multimodal data for resilient and accurate perception. %ReliFusion sets a new benchmark in reliability-driven multimodal fusion, paving the way for robust object detection in complex real-world environments. 
Future work will explore scalability to additional sensor modalities and further optimization for real-time deployment.

\newpage
\bibliographystyle{IEEEtran}
\bibliography{main}

\end{document}